\documentclass{article}

\usepackage{arxiv}
\usepackage{float}

\usepackage[utf8]{inputenc} 
\usepackage[T1]{fontenc}    
\usepackage{hyperref}       
\usepackage{url}            
\usepackage{booktabs}       
\usepackage{amsfonts}       
\usepackage{nicefrac}       
\usepackage{microtype}      
\usepackage{lipsum}

\usepackage{graphicx}
\usepackage{multirow}
\usepackage[table]{xcolor}

\title{A Deep Convolutional Network for Seismic
Shot-Gather Image Quality Classification}

\author{
  Eduardo Betine Bucker \\
  Department of Informatics\\
  PUC-Rio\\
  Rio de Janeiro, Brazil \\
  \texttt{earaujo@inf.puc-rio.br} \\
   \And
  Antonio Jos\'e~G.~Busson \\
  Department of Informatics\\
  PUC-Rio\\
  Rio de Janeiro, Brazil \\
  \texttt{busson@telemidia.inf.puc-rio.br} \\
   \And
  Ruy~Luiz~Milidi\'{u} \\
  Department of Informatics\\
  PUC-Rio\\
  Rio de Janeiro, Brazil \\
  \texttt{milidiu@inf.puc-rio.br} \\
   \And
  S\'ergio~Colcher \\
  Department of Informatics\\
  PUC-Rio\\
  Rio de Janeiro, Brazil \\
  \texttt{colcher@inf.puc-rio.br} \\
   \And
  Bruno~Pereira~Dias \\
  CENPES \\
  Petrobras\\
  Rio de Janeiro, Brazil \\
  \texttt{bpdias@gmail.com} \\
     \And
  Andr\'e~Bulc\~ao \\
  CENPES \\
  Petrobras\\
  Rio de Janeiro, Brazil \\
  \texttt{bulcao@petrobras.com.br} \\
}

\newcommand{\fscore}{94.91\%}
\newcommand{\fscoretest}{93.56\%}

\begin{document}
\maketitle

\begin{abstract}
Deep Learning-based models such as Convolutional Neural Networks, have led to significant advancements in several areas of computing applications. Seismogram quality assurance is a relevant Geophysics task, since in the early stages of seismic processing, we are required to identify and fix noisy sail lines. In this work, we introduce a real-world seismogram quality classification dataset based on 6,613 examples, manually labeled by human experts as \emph{good}, \emph{bad} or \emph{ugly}, according to their noise intensity. This dataset is used to train a CNN classifier for seismic shot-gathers quality prediction. In our empirical evaluation, we observe an F1-score of \fscoretest~in the test set.
\end{abstract}


\section{Introduction}

A classic challenge in the field of geophysics involves accurately estimating the characteristics of the Earth's subsurface based on measurements acquired by sensors on the surface. Seismic reflection is one of the most widely used methods. It involves generating seismic waves using controlled active sources on the surface (e.g., dynamite explosions in land acquisition or air guns in marine acquisition), and further collecting the reflected data with sensors located above the area~\cite{duarte2014seismic}. The term \textit{shot} refers to a firing by one of these sources. By grouping the seismic signals resulting from the same shot and registered by the sensors into a common-shot domain called the \textit{shot gather} makes it possible to produce an image that represents information about that Earth's subsurface area~\cite{yilmaz2001seismic}.
 
 Seismic shot-gather quality classification is relevant for the early stages of seismic processing and noise removal. However, controlling seismic quality is challenging as it is traditionally very subjective and time-consuming and relies upon the valuable time from skilled and experienced experts. Thus, the application of machine learning for this problem is useful as it reduces the turnaround time by quickly identifying the bad quality shot lines, instead of visual and cumbersome quality control techniques.
 
 Some previous work investigate applications of neural networks for geophysical signal classification.
 Valentine and Woodhouse \cite{valentine2010approaches},
 for instance, perform an automatic selection of high-quality seismic data to improve results from tomographic inversion.
 To accomplish that, they train an Artificial Neural Network (ANN) to recognize the frequency-domain characteristics of high and low-quality waveform. We share the same goal of selecting seismic data aimed by quality, but instead of classifying the waveforms individually,
 we take advantage of the typical shot-gather structure and build a classifier that makes an overall decision based on the correlated information presented between multiple traces.
 Jain \textit{et al.}~\cite{similarity2019} utilize a Convolutional Neural Network (CNN) to perform a similarity-based classification \cite{similarity2019, McBrearty2019}. They define an objective similarity function based on the Triplet Network \cite{hoffer2014deep},
which is a deep learning technique with widespread use in computer vision for the face recognition task.
In order to apply this procedure, the authors build a Temporal Convolutional Network (TCN) and measure pairwise similarities between the seismograms using the Triplet loss function. Further, they evaluate their approach using the Receiver Operating Characteristic Curve and report 87\% as the area under the curve (AUC) in K22A, which is the most active station in the test set from the database of USArray\footnote{http://www.usarray.org/}.

The main contributions of our work are twofold: (1) Describing the construction of a dataset for shot-gather image quality classification; (2) Presenting a comparative study of three deep learning-based approaches. The first one, using \textit{state-of-the-art} CNNs for feature extraction combined that are fed to a SVM~\cite{boser1992training} classifier. The second one, introducing our CNN architecture for shot-gather image quality classification. And the last one, fine-tuning the previous CNN. Our proposed CNN architecture achieves the best results with \fscore of F1-Score in a 10-fold cross-validation experiment.      

 The outline of this paper is structured as follows.
 Section \ref{sec:dataset} presents our dataset construction.
 Section \ref{sec:method} describe our proposed network.
 The experimentation and result analysis are presented in Section \ref{sec:experiments}. And finally, Section \ref{sec:conclusion} brings our final considerations.

\section{Seismic shot-gather quality dataset}
\label{sec:dataset}

Our dataset consists of an offshore towed streamer data in a targeted region consisting $7,993$ shot-gathers with 8 cables each, thus containing a total of $63,994$ shot-gather images. 

In a common shot-gather,
the abscissa stands for the position of the sensor relatively to the shot position,
whit this displacement being known as the offset distance.
The ordinate represents the registered time of the signal,
    the larger this time,
    the deeper the signal reached the underground surface.


\begin{figure}[H]
    \centering
    \includegraphics[width=5.0in]{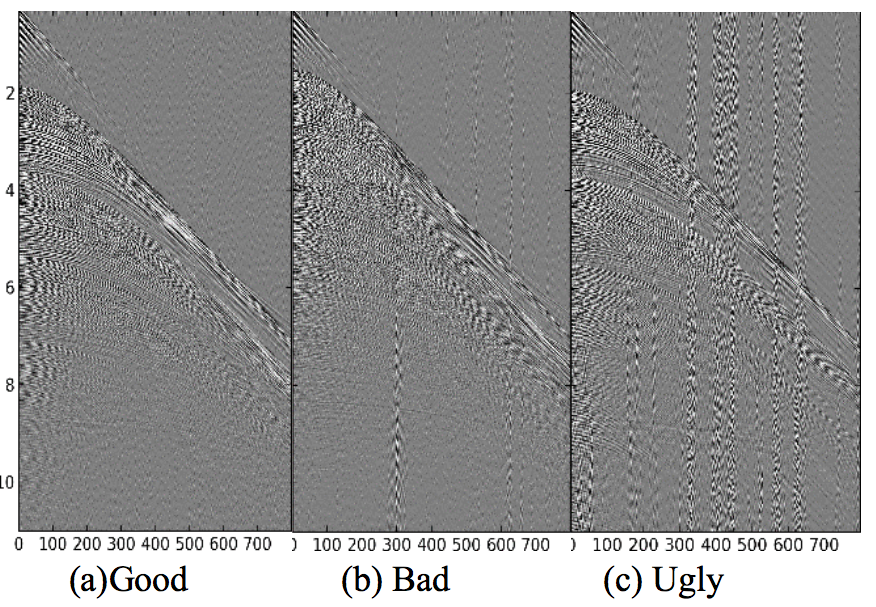}
    \caption{Examples of shot-gather images classified by a geophysicist as \emph{good}, \emph{bad} and \emph{ugly}.}
    \label{fig:seismic_classification}
\end{figure}

Out of the total generated images, $6,613$ were chosen and manually classified by a geophysicist,
    using \emph{good}, \emph{bad} and \emph{ugly} labels,
    according to a visual inspection of artifacts related to swell noise and anomalous recorded amplitude.
This process takes roughly ten hours of human labor.
Figure~\ref{fig:seismic_classification} shows
    one example per class of shot-gather images classified by a geophysicist.
From this figure,
we observe that
    the \emph{good} label represents clean images,
    while the \emph{ugly} label represents images that have an intense presence of noise.
    The \emph{bad} label represents those images who are neither fully clean, nor as noisy as the \emph{ugly} class.

The table~\ref{tab:dataset_distribution} shows the shot-gather quantity and proportion of the three labels. The dataset is more represented by \emph{good} (66.68\%), followed by the \emph{bad} (31.54\%) and the \emph{ugly} (1.76\%). This indicates that we are working with an acquisition, where \emph{good} and \emph{bad} labels almost dominate the entire representation. This dataset class imbalance is very challenging to machine learning algorithms, since they are data-intensive and we have just a few \emph{ugly} shot-gather images. The Train represents 79\% and Test 21\%.

\begin{table}[h]
    \centering
    \caption{Shot-gather images distribution over the three categories}
    \begin{tabular}{|r| r r r |r c|}
        \cline{2-6}
        \multicolumn{1}{c|}{} & 
        \cellcolor{gray!50} \textbf{Good} & \cellcolor{gray!50}\textbf{Bad} & \cellcolor{gray!50} \textbf{Ugly} & \cellcolor{gray!50} \textbf{Total} &
        \cellcolor{gray!50} \textbf{(\%)}\\
        \hline
        \cellcolor{gray!50} \textbf{Train} & 3,364 & 1,746 & 114 & 5,224 & 79.0 \\
        \cellcolor{gray!50} \textbf{Test} & 1,064 & 340 & 3 & 1,389 & 21.0 \\
        \hline
        \cellcolor{gray!50} \textbf{Total} & 4,410 & 2,086 & 117 & 6,613 & 100 \\
        \cellcolor{gray!50} \textbf{(\%)} & 66.68 & 31.54 & 1.76 & 100 & \\
        \hline
    \end{tabular}
    \label{tab:dataset_distribution}
\end{table}

\section{Minception Network}
\label{sec:method}

\begin{figure} [H]
    \centering
    \includegraphics[width=0.3\textwidth]{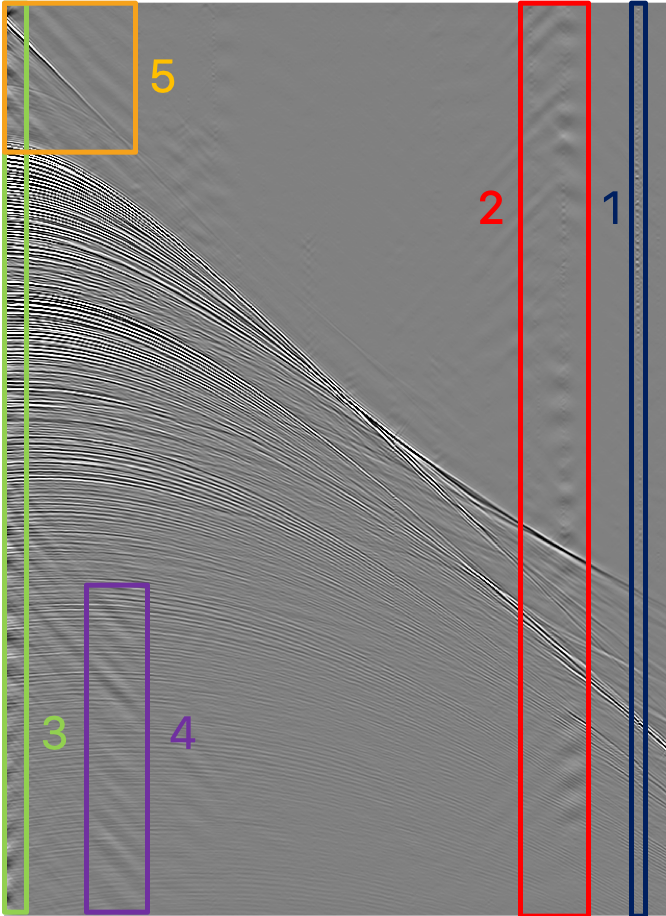}
    \caption{Example of noise variations in seismic marine survey made by Elboth et al. \cite{elboth2009attenuation}: (1) Pressure Variation Noise;  (2) Swell Noise; (3) Tugging/Strumming Noise; (4) Seismic Interference  and (5) Proppeller Cavitation Noise.}
    \label{fig:sample_noises}
\end{figure}

 A typical CNN usually has a stack of convolutional layers with 3x3 or 5x5 kernels followed by a pooling layer. Due to the variance of shot-gather noise shape and scale, select an appropriate kernel size is a key challenge. The shot-gather image presented in Figure~\ref{fig:sample_noises} shows examples of noise variations acquired through a seismic marine survey \cite{elboth2009attenuation}. To detect larger noises, higher kernel sizes are more adequate, while to thin noises, small kernel sizes work better. 

 Inspired on GoogleNet's Inception block~\cite{szegedy2015going}, we suggest the use of both 3x3 and 5x5 kernel sizes at the same layer. In this way, during networking training, the internal layers automatically choose the kernel sizes that will be relevant to learn the required information. 

\begin{figure}[H]
    \centering
    \includegraphics[width=\textwidth]{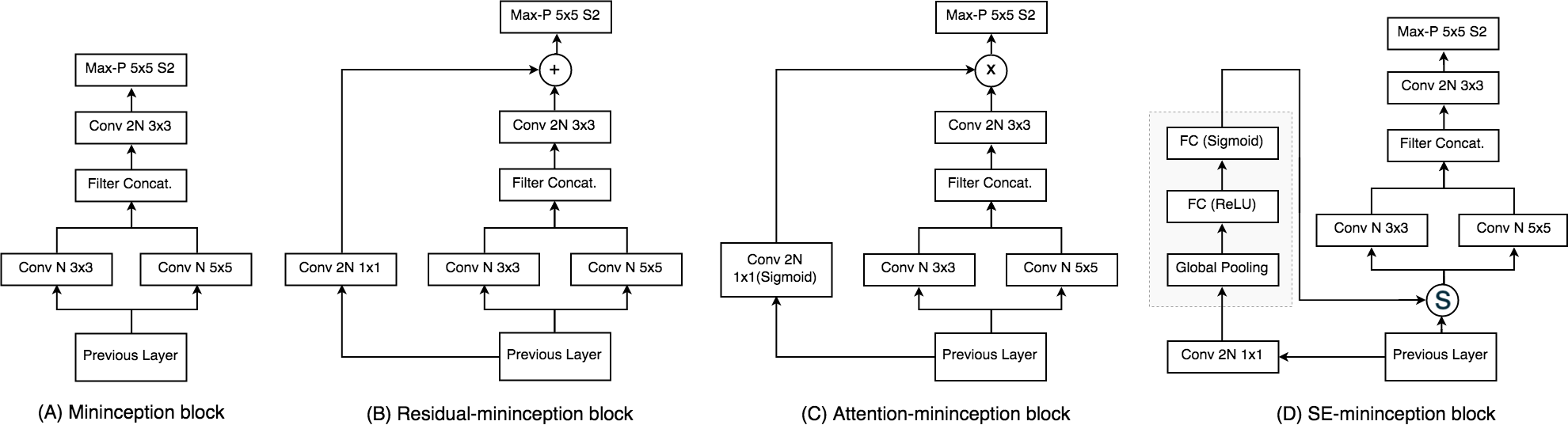}
    \caption{(A) Standard minception block. (B) Minception block with residual connection. (C) Minception block with self attention mechanism. (D) Minception combined with Squeeze-and-Excitation (SE) block. }
    \label{fig:miniception_blocks}
\end{figure}

The Figure~\ref{fig:miniception_blocks} (A) shows the standard Minception block, a simplified version of Inception block with fewer parameters. Instead of using a single convolutional layer, we combine a 3x3 and 5x5 convolutions by concatenating their output in a single tensor forming the input of a 3x3 convolution that creates new filters by correlating the previous one. 

Table~\ref{tab:minception_net} shows the Minception network structure, that is built on four stacked Minception blocks. All convolutional layers use ReLU nonlinearity, same padding, and stride 1. The $N$ hyperparameter indicates the number of kernels per layer, respectively equals to  1,  2,  4  and  8. A final downsampling is handled with a 5x5 max pooling with stride 2 to perform spatial dimensionality reduction.  The last  two  layers,  which  are  fully  connected,  has  32  units  and feeds a softmax layer with 3 classes.

\begin{table}[H]
    \centering
    \caption{MinceptionNet architecture}
    \begin{tabular}{|l|l|}
         \hline
         \cellcolor{gray!50} \textbf{Component} & \cellcolor{gray!50} \textbf{output shape} \\
         \hline
         Input image & 299x299x1 \\
         Down-Sampling (Area interpolation) & 99x99x1 \\
         Minception block 1 (N=1) & 97x97x2 \\
         Max Pooling  & 47x47x2 \\
         Minception block 2 (N=2) & 45x45x4 \\
         Max Pooling  & 21x21x4 \\
         Minception block 3 (N=4) & 19x19x8 \\
         Max Pooling  & 8x8x8 \\
         Minception block 4 (N=8) & 6x6x16 \\
         Max Pooling  & 1x1x16 \\
         Fully connected (ReLU)  & 32\\
         Fully connected (Linear)& 32 \\
         Softmax & 3 \\
         \hline
    \end{tabular}
    \label{tab:minception_net}
\end{table}

We explored the other three variants of minception blocks using mechanisms of benchmarks networks for image classification. The first one, called (B) Residual-Minception  uses the mechanism proposed by ResNet Network that has a goal to address the vanishing gradient problem of Deep Neural Networks. As described in \cite{He2016DeepRL} it's empirically known that when we increase the depth of the network, the gradient gets saturated and then degrades rapidly. To solve it we want to experiment with the addition of a skip-connection to Standard Minception. However, to sum it properly, we need to previously add a convolution block 2N 1x1 in the skip-connection just to match with the shape from the output of Conv 2N 3x3.

The second variant, (C) Attention-Minception uses the mechanism of self-attention \cite{Woo2018CBAMCB}. This mechanism is similar to the previous block as it has a skip-connection with a Convolutional Block, but in this particular case we use the sigmoid activation function instead of the standard ReLU, then it performed an \textit{element-wise} multiplication.

The last Minception variant, (D) SE-Minception incorporates the channel attention mechanism provided by Squeeze and Excitation Module (SE) from SE-Net~\cite{Hu18}. Squeeze and Excitation is a technique that improves the quality of representations produced by the network by learning global information from channels and dynamically emphasizing informative features. The process of SE is divided into three operations: (1) squeeze spatial information into a channel descriptor, (2) capture channel-wise dependencies targeting on important features and (3) perform a scale operation (channel-wise multiplication) in the original channels. The first operation is performed by a Global Average Pooling, which generates a summary of channel-wise statistics. The second operation is performed by two fully-connected (FC) layers in a row, where the first layer compress channel dimensionality by a given reduction ratio and the second layer increase channel dimensions back to the input channel size.
\section{Experimental Evaluation}
\label{sec:experiments}

In this section, we evaluate the effectiveness of proposed approaches for seismic shot-gather image quality classification. First, we report the performance of the SVM classifier combined with transfer learning method. Next, we investigate the performance of our end-to-end CNN called MinceptionNet. And finally, we optimize the MinceptionNet exploring the trade-offs by tunning the network parameters based on two hyper-parameters, a width multiplier and a resolution multiplier.

\subsection{Setup}
\label{subsec:experiment_setup}

We use Adam as the default optimizer
and set
    the learning rate to 0.001,
    the batch size to 64,
    the training phase to 100 epochs with patience at 10 epochs,
        looking always for the best accuracy in the validation set.
 
 The Dataset has imbalanced classes, whit the \emph{ugly} label barely hitting 2\% of the total. We perform a 10-fold cross-validation and evaluate the model by the F1-Score per class and the F1-weighted (F1-W), which is the sum of the F1-score for each class weighted by its proportion.

\begin{equation}
\label{equation:precision}
P_l = \frac{TP_l}{TP_l + FP_l}
\end{equation}

\begin{equation}
\label{equation:recall}
R_l = \frac{TP_l}{TP_l + FN_l}
\end{equation}

\begin{equation}
\label{equation:f1}
F1_l = \frac{2 \times P(l) \times R_l}{P_l + R_l}
\end{equation}

\begin{equation}
\label{equation:f1-weighted}
F1_{weigthed} = \sum_{l}^{|L|} {F1_l \times p_l }
\end{equation}

Where $TP_l, TN_l, FP_l$, $FN_l$ and $p_l$  denote the examples that are true positives, true negatives, false positives, false negatives, and proportion of label $l$, respectively. The F1 score, defined in Equation~\ref{equation:f1}, measures how precise the classifier by the harmonic mean between Precision (Equation~\ref{equation:precision}) and Recall (Equation~\ref{equation:recall}).
 
\subsection{Baseline}

\begin{figure*}
    \centering
    \includegraphics[width=\textwidth]{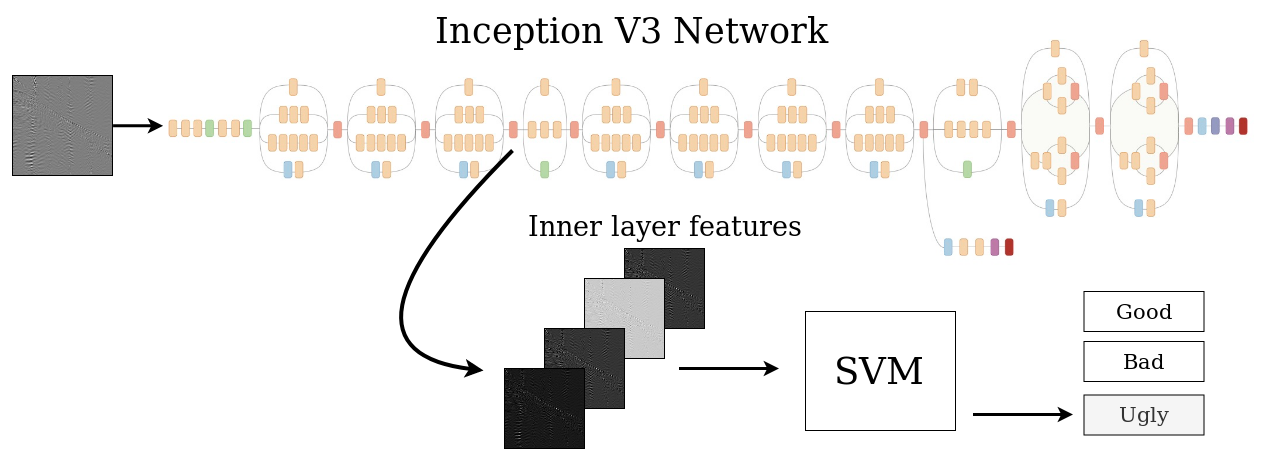}
    \caption{Workflow of our baseline. Given a shot-gather image, features extracted from intermediate convolutional layers. Then SVM is trained to predict the classification output.}
    \label{fig:transf_learning_workflow}
\end{figure*}

For the baseline method, we use the SVM~\cite{boser1992training} classifier with input features provided by transfer learning.
We use the output of a chosen hidden layer of a pre-trained CNN for the object recognition task. We use the inner layers of the VGG16, VGG19~\cite{simonyan2014very} and InceptionV3~\cite{szegedy2016rethinking} with the network weights trained in the ImageNet classification dataset~\cite{deng2009imagenet}.
The workflow of this approach using the InceptionV3 network is illustrated in Figure~\ref{fig:transf_learning_workflow}.

CNNs when trained tend to learn at the first layers features that resemble either Gabor filters or color blobs. At the intermediate and final layers, the combination of these filters helps to extract relevant features from images, resulting in complex patterns~\cite{yosinski2015understanding}. Once we have extracted features from each layer,  we then compare the performances of SVM classifier using each of them.

Tables \ref{tab:exp1_vgg_16_results}, \ref{tab:exp1_vgg_19_results} and \ref{tab:exp1_inception_results} summarize the results for VGG16, VGG19, and InceptionV3, respectively. The best model uses features extracted from VGG19's block 2, which produces an F1-W of 90.82\%. Next, VGG16's block 2 is the second best model, achieving an F1-W of 88.99\%. We can see that the Block 2 is the best for both networks, where we also get the best F1-Good and F1-Bad. Analysing VGG16 and VGG19 results, we observe that both models have similar results, but VGG19 is slightly better. InceptionV3 is our worst performing model, since it does not reach higher results than VGG19 or VGG16.

\begin{table}[H]
    \centering
    \caption{F1-W and F1-score for Good (G), Bad (B) and Ugly (U) labels with SVM over the VGG16's blocks}
    \begin{tabular}{|c|r|r|r|r|}
         \hline
         \cellcolor{gray!50} \textbf{Block} & \cellcolor{gray!50} \cellcolor{gray!50}\textbf{F1-W (\%)} & \cellcolor{gray!50} \textbf{F1-G (\%)} & \cellcolor{gray!50} \textbf{F1-B (\%)} &  \cellcolor{gray!50} \textbf{F1-U (\%)}  \\
         \hline
         \multirow{2}{*}{\footnotesize{0}} & 83.03 & 91.8 & 66.58 & 45.95  \\
         & $\pm$ 0.45 & $\pm$ 0.09 & $\pm$ 0.29 & $\pm$ 0.93 \\
         \hline
         
          \multirow{2}{*}{\footnotesize{1}} & 88.37 & 94.20 & 78.21 & 50.02  \\
         & $\pm$ 0.24 & $\pm$ 0.16 & $\pm$ 0.42 & $\pm$ 1.34 \\
         \hline
         
          \multirow{2}{*}{\footnotesize{2}} & \textbf{88.99} & \textbf{95.08} & \textbf{81.48} & 49.96  \\
         & $\pm$ 0.30 & $\pm$ 0.13 & $\pm$ 0.35 & $\pm$ 0.99 \\
         \hline
         
          \multirow{2}{*}{\footnotesize{3}} & 74.93 & 87.99 & 48.71 &  \textbf{50.61}  \\
         & $\pm$ 0.46 & $\pm$ 0.05 & $\pm$ 0.38 & $\pm$ 1.36 \\
         \hline
         
          \multirow{2}{*}{\footnotesize{4}} & 36.51 & 47.98 & 14.39 & 0.36  \\
         & $\pm$ 6.62 & $\pm$ 12.39 & $\pm$ 6.95 & $\pm$ 0.34 \\
        \hline
    \end{tabular}
    \label{tab:exp1_vgg_16_results}
\end{table}

\begin{table}[H]
    \centering
    \caption{F1-W and F1-score for \emph{good} (G), \emph{bad} (B) and \emph{ugly} (U) labels with SVM over the VGG19's blocks}
    \begin{tabular}{|c|r|r|r|r|}
         \hline
         \cellcolor{gray!50} \textbf{Block} & \cellcolor{gray!50} \textbf{F1-W (\%)} & \cellcolor{gray!50} \textbf{F1-G (\%)} & \cellcolor{gray!50} \textbf{F1-B (\%)} & \cellcolor{gray!50} \textbf{F1-U (\%)}  \\
         \hline
         \multirow{2}{*}{\footnotesize{0}} & 83.06 & 91.86 & 66.58 & 45.46  \\
         & $\pm$ 0.46 & $\pm$ 0.08 & $\pm$ 0.28 & $\pm$ 0.86 \\
         \hline
          \multirow{2}{*}{\footnotesize{1}} & 87.76 & 93.94 & 76.87 & 48.78  \\
         & $\pm$ 0.38 & $\pm$ 0.21 & $\pm$ 0.67 & $\pm$ 1.01 \\
         \hline
          \multirow{2}{*}{\footnotesize{2}} & \textbf{90.82} & \textbf{95.64} & \textbf{83.10} & 47.21  \\
         & $\pm$ 0.29 & $\pm$ 0.18 & $\pm$ 0.44 & $\pm$ 1.53 \\
         \hline
          \multirow{2}{*}{\footnotesize{3}} & 81.49 & 90.8 & 63.59 &  \textbf{50.14}  \\
         & $\pm$ 0.58 & $\pm$ 0.13 & $\pm$ 0.64 & $\pm$ 0.88 \\
         \hline
          \multirow{2}{*}{\footnotesize{4}} & 36.51 & 47.98 & 14.39 & 0.36  \\
         & $\pm$ 6.62 & $\pm$ 12.39 & $\pm$ 6.95 & $\pm$ 0.34 \\
        \hline
    \end{tabular}
    \label{tab:exp1_vgg_19_results}
\end{table}

\begin{table}[H]
    \centering
    \caption{F1-W and F1-score for \emph{good} (G), \emph{bad} (B) and \emph{ugly} (U) labels with SVM over the InceptionV3's blocks}
    \begin{tabular}{|c|r|r|r|r|}
         \hline
         \cellcolor{gray!50} \textbf{Block} & \cellcolor{gray!50} \textbf{F1-W (\%)} & \cellcolor{gray!50} \textbf{F1-G (\%)} &  \cellcolor{gray!50} \textbf{F1-B (\%)} & \cellcolor{gray!50} \textbf{F1-U (\%)} \\
         \hline
         \multirow{2}{*}{\footnotesize{0}}  & \textbf{86.02} & \textbf{93.23} & \textbf{73.07} & \textbf{45.56}  \\
         & $\pm$ 0.30 & $\pm$ 0.17 & $\pm$ 0.55 & $\pm$ 1.45 \\
         \hline
          \multirow{2}{*}{\footnotesize{1}}  & 85.54 & 92.99 & 72.05 & 45.40  \\
         & $\pm$ 0.40 & $\pm$ 0.17 & $\pm$ 0.60 & $\pm$ 1.10 \\
         \hline
          \multirow{2}{*}{\footnotesize{2}} & 81.01 & 90.89 & 62.13 & 45.22 \\
         & $\pm$ 0.52 & $\pm$ 0.08 & $\pm$ 0.53 & $\pm$ 1.3 \\
         \hline
          \multirow{2}{*}{\footnotesize{3}} & 83.21 & 91.98 & 66.86 & 44.35  \\
         & $\pm$ 0.38 & $\pm$ 0.14 & $\pm$ 0.56 & $\pm$ 1.14 \\
         \hline
          \multirow{2}{*}{\footnotesize{4}} & 81.5 & 91.23 & 63.04 & 43.77 \\
         & $\pm$ 0.45 & $\pm$ 0.14 & $\pm$ 0.59 & $\pm$ 1.05 \\
         \hline
         \multirow{2}{*}{\footnotesize{5}} & 81.09 & 90.98 & 62.25 & 44.36 \\
         & $\pm$ 0.39 & $\pm$ 0.11 & $\pm$ 0.58 & $\pm$ 1.52 \\
         \hline
         \multirow{2}{*}{\footnotesize{6}} & 77.92 & 89.63 & 55.07 & 44.19 \\
         & $\pm$ 0.51 & $\pm$ 0.04 & $\pm$ 0.47 & $\pm$ 1.51 \\
         \hline
         \multirow{2}{*}{\footnotesize{7}} & 76.78 & 89.16 & 52.41 & 44.59  \\
         & $\pm$ 0.49 & $\pm$ 0.0 & $\pm$ 0.40 & $\pm$ 1.50 \\
         \hline
         \multirow{2}{*}{\footnotesize{8}}  & 83.05 & 91.99 & 66.36 & 43.58 \\
         & $\pm$ 0.41 & $\pm$ 0.11 & $\pm$ 0.48 & $\pm$ 1.19 \\
         \hline
         \multirow{2}{*}{\footnotesize{9}}  & 76.10 & 89.06 & 50.64 & 41.46 \\
         & $\pm$ 0.53 & $\pm$ 0.0 & $\pm$ 0.44 & $\pm$ 1.32 \\
         \hline
         \multirow{2}{*}{\footnotesize{10}} & 84.42 & 92.73 & 69.22 & 42.68 \\
         & $\pm$ 0.38 & $\pm$ 0.12 & $\pm$ 0.51 & $\pm$ 1.19 \\
         \hline
    \end{tabular}
    \label{tab:exp1_inception_results}
\end{table}

\subsection{Minception Network}
\label{subsec:minception_models}

Table~\ref{tab:exp1_results} shows the results for each MinceptionNet variation.
The best model is obtained with the architecture SE-MiniceptionNet (D),
which produces an F1-W of 93.84\%.
The Architecture Residual MinceptionNet (B) is the second best model,
producing a F1-W of 93.76\%.
Our basic architecture Standard Minception (A) is the third best model,
achieving a F1-W of 92.90\%.
The architecture Attention MinceptionNet (C), achieves a F1-W of only 91.09\%.
The SE-MinceptionNet also reaches
    the best F1-Good and F1-Ugly
among other MinceptionNet variations.

\begin{table}[!hbt]
    \centering
    \caption{Comparison of F1-W and F1-score for \emph{good} (G), \emph{bad} (B) and \emph{ugly} (U) between MinceptionNet versions}
    \begin{tabular}{|l|r|r|r|r|}
         \hline
         \cellcolor{gray!50} \textbf{Model} & \cellcolor{gray!50} \textbf{F1-W (\%)} & \cellcolor{gray!50} \textbf{F1-G (\%)} &  \cellcolor{gray!50} \textbf{F1-B (\%)} & \cellcolor{gray!50} \textbf{F1-U (\%)}  \\
         \hline
         
         \multirow{2}{*}{\footnotesize{SE-MinceptionNet (D)}} & \textbf{93.84} & \textbf{96.58} & 90.51 & \textbf{50.26} \\
         & $\pm$ 0.25 & $\pm$ 0.24 & $\pm$ 0.55 & $\pm$ 7.24 \\
         \hline
         
         \multirow{2}{*}{\footnotesize{Attention MinceptionNet (C)}} & 91.09 & 95.2 & 85.27 & 40.13 \\
         & $\pm$ 1.71 & $\pm$ 0.73 & $\pm$ 1.28 & $\pm$ 7.72 \\
         \hline
         
         \multirow{2}{*}{\footnotesize{Residual MinceptionNet (B)} } & 93.76 & 96.55 & \textbf{90.52} & 46.88 \\
         & $\pm$ 0.21 & $\pm$ 0.29 & $\pm$ 0.55 & $\pm$ 7.14 \\
         \hline
         
         \multirow{2}{*}{\footnotesize{Standard MinceptionNet (A)}} & 92.90 & 96.00 & 89.20 & 42.60 \\
         & $\pm$ 0.39 & $\pm$ 0.45 & $\pm$ 0.88 & $\pm$ 9.19 \\

     \hline
    \end{tabular}
    \label{tab:exp1_results}
\end{table}
\vfill

\subsection{Minception Tunning}
\label{subsec:minception_models}

In this phase, we optimize the best version of the MinceptionNet (SE-MinceptionNet) architecture making some fine adjusts in our initial network.
Hence,
we perform a search on the kernel quantity multiplier alpha.
When alpha is equal to one,
that is exactly the default SE-MinceptionNet,
but as soon as alpha increases,
we multiply the initial number of filters in each convolution per alpha.
Therefore,
we expect that increasing the number of filters per stage and the depth of the network,
we should probably boost the network performance.
Nevertheless,
increasing these hyper-parameters may cause a strong impact in the computational effort and consequently in the computational budget.
Hence,
we have a trade-off between optimization and computational budget.

To search for the best parameter configuration,
we perform a grid search on the kernel quantity multiplier
    $\alpha \in \{1,2,3,4,5,6,7,8,9,10\}$.
Table~\ref{tab:exp2_gridsearch} shows
    the results of SE-MinceptionNet for each $\alpha$ value.
The best $\alpha$ value is 8,
which produces a F1-W of \fscore, F1-G of 97.27\%, F1-B of 92.08\% and F1-U of 56.01\%. 

\begin{table}[H]
\centering
\caption{SE-MinceptionNet F1-W and F1-score for \emph{good} (G), \emph{bad} (B) and \emph{ugly} (U) with different $\alpha$ multiplier of network's kernels}
\begin{tabular}{|c|c| c|c|c|c| }
\cline{3-6}
\multicolumn{2}{c|}{} & \cellcolor{gray!50} \textbf{F1-W (\%)} & \cellcolor{gray!50}  \textbf{F1-G (\%)} & \cellcolor{gray!50} \textbf{F1-B (\%)} & \cellcolor{gray!50} \textbf{F1-U (\%)} \\
\hline
    \multirow{2}{*} & \textbf{1} & 93.84  $\pm$ 0.25 & 96.58  $\pm$ 0.24 & 90.51  $\pm$ 0.55 & 50.26  $\pm$ 7.24 \\
    
    \multirow{2}{*} & \textbf{2} & 94.11  $\pm$ 0.21 & 96.68 $\pm$ 0.29 & 90.82 $\pm$ 0.57 & 55.72 $\pm$ 4.86 \\
    
    \multirow{2}{*} & \textbf{3} & 94.38  $\pm$ 0.26 & 96.88 $\pm$ 0.30 & 91.29 $\pm$ 0.65 & 55.15 $\pm$ 5.19 \\
    
    \multirow{2}{*} & \textbf{4} & 94.42  $\pm$ 0.32 & 96.94 $\pm$ 0.39 & 91.32 $\pm$ 0.85 & 54.83 $\pm$ 5.35 \\
    
    \multirow{2}{*}{\textbf{$\alpha$}} & \textbf{5} & 94.37  $\pm$ 0.34 & 96.88 $\pm$ 0.26 & 91.20 $\pm$ 0.60 & 56.64 $\pm$ 5.84 \\
    
    \multirow{2}{*} & \textbf{6} & 94.60  $\pm$ 0.25 & 97.09 $\pm$ 0.24 & 91.73 $\pm$ 0.43 & 51.92 $\pm$ 6.17 \\
    
    \multirow{2}{*} & \textbf{7} & 94.61  $\pm$ 0.36 & 97.14 $\pm$	0.31 &	91.71 $\pm$	0.67 &	51.15 $\pm$	8.10 \\
    
    \multirow{2}{*} & \textbf{8} & \textbf{94.91  $\pm$ 0.29} & \textbf{97.27 $\pm$	0.21} &	\textbf{92.08 $\pm$	0.51} &	\textbf{56.01 $\pm$	5.59} \\
    
    \multirow{2}{*} & \textbf{9} & 94.52  $\pm$ 0.33 & 97.16 $\pm$	0.34 &	91.34 $\pm$	0.95 &	51.80 $\pm$	7.09 \\
    
    \multirow{2}{*} & \textbf{10} & 94.59  $\pm$ 0.21 & 97.05 $\pm$	0.29 &	91.62 $\pm$	0.75 &	54.70 $\pm$	5.49 \\
    
\hline
\end{tabular}
\label{tab:exp2_gridsearch}
\end{table}

\subsection{Model Results}
\label{subsec:summary_table}

In Table~\ref{tab:exp1_results}, we summarize our empirical findings in the five examined models. The best model was obtained with SE-MinceptionNet with $\alpha = 8$, which produced a F1-W of 94.91\%. Standard SE-MinceptionNet ($\alpha = 1$) was the second best model, producing a F1-W of 93.84\%. Next, our baselines VGG19, VGG16, and InceptionV3 were the third, fourth, and fifth place, achieving a F1-W of 90.82\%, 88.99\%, and 86.02\%, respectively.

\begin{table} [H]
    \centering
    \caption{Summary Comparison of F1-W and F1-score for \emph{good} (G), \emph{bad} (B) and \emph{ugly} (U) between models}
    \begin{tabular}{|l|r|r|r|r|}
         \hline
         \cellcolor{gray!50} \textbf{Model} & \cellcolor{gray!50} \textbf{F1-W (\%)} & \cellcolor{gray!50} \textbf{F1-G (\%)} &  \cellcolor{gray!50} \textbf{F1-B (\%)} & \cellcolor{gray!50} \textbf{F1-U (\%)} \\
         \hline
         
         \multirow{2}{*}{\footnotesize{\textbf{SE-MinceptionNet*8}}} & \textbf{94.91} & \textbf{97.27} & \textbf{92.08} & \textbf{56.01}  \\
         & $\pm$ 0.29 & $\pm$ 0.21 & $\pm$ 0.51 & $\pm$ 5.59  \\
         \hline
         
         \multirow{2}{*}{\footnotesize{SE-MinceptionNet}} & 93.84 & 96.58 & 90.51 & 50.26  \\
         & $\pm$ 0.25 & $\pm$ 0.24 & $\pm$ 0.55 & $\pm$ 7.24  \\
         \hline
         \hline
         
         \multirow{2}{*}{\footnotesize{VGG19's block 2 + SVM}} & 90.82 & 95.64 & 83.10 & 47.21  \\
         & $\pm$ 0.29 & $\pm$ 0.18 & $\pm$ 0.44 & $\pm$ 1.53  \\
         \hline
         
         \multirow{2}{*}{\footnotesize{VGG16's block 2 + SVM}} & 88.99 & 95.08 & 81.48 & 49.96 \\
         & $\pm$ 0.30 & $\pm$ 0.13 & $\pm$ 0.35 & $\pm$ 0.99  \\
         \hline
         
         \multirow{2}{*}{\footnotesize{InceptionV3's block 0 + SVM}} & 86.02 & 93.23 & 73.07 & 45.56 \\
         & $\pm$ 0.30 & $\pm$ 0.17 & $\pm$ 0.55 & $\pm$ 1.45  \\
         \hline
        
    \end{tabular}
    \label{tab:exp1_results}
\end{table}

\clearpage
In Table~\ref{tab:exp1_results_test}, we show the results of SE-MinceptionNet with $\alpha = 8$ in the test set, where it achieves an F1-W of \fscoretest, F1-G of 95.94\%, F1-B of 86.80\% and F1-U of 28.57\%. The model presents better result for F1-G in comparison with the train set and also show high recalls values, producing a Recall-G of 92.73\%, Recall-B of 93.82\% and Recall-U of 100\%. In contrast, we notice a decrease in F1-B and F1-U, due to Precision-B of 80.76\% and Precision-U of 16.67\%.


\begin{table}[H]
    \centering
    \caption{F1-W, F1, Recall and Precision for \emph{good} (G), \emph{bad} (B) and \emph{ugly} (U) of SE-MinceptionNet with $\alpha = 8$ in test set}
    \begin{tabular}{|r|c|c|c|}
         \hline
         \cellcolor{gray!50} \textbf{F1-W (\%)} &  \multicolumn{3}{c|}{\textbf{93.56}} \\
         \hline
         \hline
         \hline
         \cellcolor{gray!50} \textbf{} &
         \cellcolor{gray!50} \textbf{F1 (\%)} &
         \cellcolor{gray!50} \textbf{Recall (\%)} &
         \cellcolor{gray!50} \textbf{Precision (\%)} \\
         \hline
        
         \hline
         \cellcolor{gray!50} \textbf{Good} & 95.94 & 92.73 & 99.39 \\ 
         \hline
         \cellcolor{gray!50} \textbf{Bad} & 86.80 & 93.82 & 80.76 \\ 
         \hline
         \cellcolor{gray!50} \textbf{Ugly} & 28.57 & 100.0 & 16.67 \\ 
         \hline
         
    \end{tabular}
    \label{tab:exp1_results_test}
\end{table}

\begin{figure}[H]
    \centering
    \includegraphics[width=\textwidth]{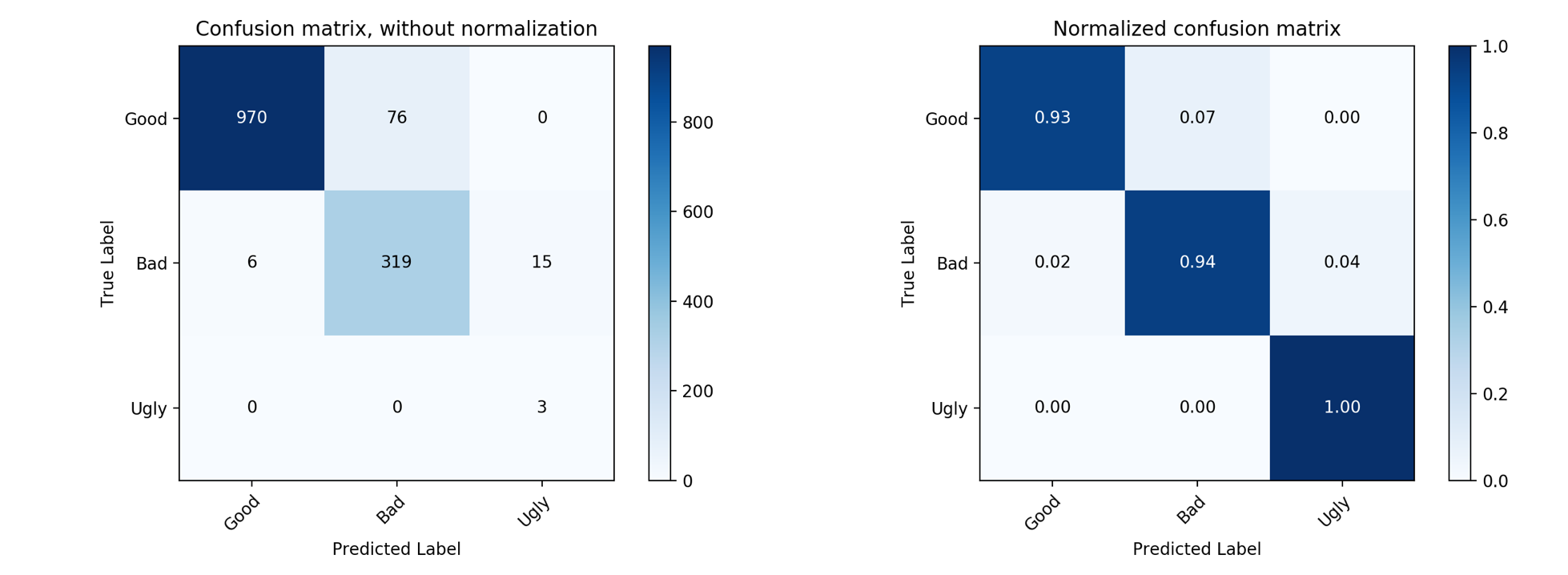}
    \caption{Left: Confusion Matrix for the test set; Right: Confusion Matrix with normalization.}
    \label{fig:confusion_matrix}
\end{figure}

Figure \ref{fig:confusion_matrix} shows the confusion matrix. The model correctly identified 970 \textit{good} images, 319 \textit{bad} images, and 3  \textit{ugly} images. We can observe that the model major confusion is between contiguous classes, where it misclassifies 76 \textit{good} images as \textit{bad} and 15 \textit{bad} images as \textit{ugly}. As strong points, the model produces high recall values and has a high precision for \textit{good}, but in regards to \textit{ugly} label, our Recall-U of 100\% is suspicious, since we have just 3 \textit{ugly} images in test set (1.76\% of total). The weak points of the model are relative to the low precision for \textit{bad} and \textit{ugly}, as we notice that the model predicted many images incorrectly for that classes. 


\section{Conclusion}
\label{sec:conclusion}

 In this work,
 we examine the application of deep learning to perform a quality classifier of shot-gather noisy images.
 For that sake,
 we build a dataset with $6,613$ seismic shot-gather images and manually classified them using
    \emph{good}, \emph{bad} and \emph{ugly} labels,
    according to a visual inspection of artifacts related to swell noise and anomalous recorded amplitude. We propose the Minception Net, a network architecture inspired on GoogleNet's Inception Block,
    but with fewer parameters. Additionally, we also propose three Minception variants that incorporate mechanisms as residual connection (Residual Minception), attention (Attention Minception) and squeeze-and-excitation (SE-Minception).
 
In a 10-fold cross validation experiment,
we use as baseline three CNNs InceptionV3, VGG16 and VGG19 to extract features from shot-gather images and apply their values as the input of an SVM classifier.
The best result is obtained by the second block from VGG19 with an F1-W of 90.82\%.
The SE-Minception variant achieves the best result with an F1-W of 93.84\% .
Then,
we optimize SE-MinceptionNet by performing a $\alpha$ search,
where $\alpha$ is a parameter that multiplies the quantity of initial kernels from the standard minception block.
With this search, we are looking for a balance between the depth and the width of the network. The validation results show that setting alpha to 8 upgrades the F1-W to \fscore. In the test set, the SE-MinceptionNet produced a final F1-W of 93.5\%. 

As future work, we plan to use deep learning models for data augmentation to increase the quantity of \emph{ugly} seismic shot-gather images in training. We expect this will produce a reliable recall and improve the value of F1-score for \emph{ugly} label.
 
\section*{Acknowledgment}

The authors would like to thank Petrobrás S.A.
    for the collaboration and
    for providing the dataset.
The authors would also thank
    Raphael Rocha, Luis Felipe Müller, Matheus Cabral, Miguel de Brito and Ivan Pereira
    from the Pontifical Catholic University of Rio de Janeiro.

\bibliographystyle{unsrt}  
\bibliography{references}  


\end{document}